%% file: main.tex
\documentclass[a4paper, 10pt, conference]{ieeeconf}      

\IEEEoverridecommandlockouts                              

\usepackage[shortcuts,acronym]{glossaries}

\newacronym{ml}{ML}{Machine Learning}
\newacronym{dl}{DL}{Deep Learning}
\newacronym{rpn}{RPN}{Region Proposal Network}
\newacronym{roi}{ROI}{Region of Interest}
\newacronym{r-cnn}{R-CNN}{Region-based Convolutional Neural Network}
\newacronym{cnn}{CNN}{Convolutional Neural Network}
\newacronym{rcs}{RCS}{Radar Cross Section}
\newacronym{ah}{AH}{Adaptive Height}
\newacronym{fh}{FH}{Fixed Height}
\newacronym{ae}{AE}{Azimuth Extension}
\newacronym{aue}{AUE}{Azimuth Uncertainty Extension}
\newacronym{swfb}{SWFB}{Self-Weighted Fusion Block}
\newacronym{safb}{SAFB}{Similarity-based Attention Fusion Block}
\newacronym{fpn}{FPN}{Feature Pyramid Network}
\newacronym{map}{mAP}{mean Average Precision}
\newacronym{iou}{IoU}{Intersection-over-Union}
\newacronym{cbam}{CBAM}{Convolutional Block Attention Module}
\usepackage{times}
\usepackage{epsfig}
\usepackage{graphicx}
\newsavebox\mybox
\usepackage{amsmath}
\usepackage{amssymb}
\usepackage{subcaption}
\usepackage{amsmath}
\usepackage{multirow}
\usepackage{float}
\usepackage[flushleft]{threeparttable}
\usepackage[font=small,labelfont=bf]{caption} 

\title{\LARGE \bf
Multi-Task Cross-Modality Attention-Fusion for 2D Object Detection
}

\author{Huawei Sun$^{1,2}$, Hao Feng$^{2}$, Georg Stettinger$^{1}$, Lorenzo Servadei$^{2}$, Robert Wille$^{2}$
\thanks{$^{1}$Infineon Technologies AG, Neubiberg, Germany
        {\tt\small \{huawei.sun, georg.stettinger\}@infineon.com}}%
\thanks{$^{2}$Technical University of Munich, Munich, Germany
        {\tt\small \{hao.feng, lorenzo.servadei, robert.wille\}@tum.de}}%
}

\begin{document}


\maketitle
\thispagestyle{empty}
\pagestyle{empty}

\begin{abstract}
Accurate and robust object detection is critical for autonomous driving. Image-based detectors face difficulties caused by low visibility in adverse weather conditions. Thus, radar-camera fusion is of particular interest but presents challenges in optimally fusing heterogeneous data sources. To approach this issue, we propose two new radar preprocessing techniques to better align radar and camera data. In addition, we introduce a Multi-Task Cross-Modality Attention-Fusion Network (MCAF-Net) for object detection, which includes two new fusion blocks. These allow for exploiting information from the feature maps more comprehensively. The proposed algorithm jointly detects objects and segments free space, which guides the model to focus on the more relevant part of the scene, namely, the occupied space. Our approach outperforms current state-of-the-art radar-camera fusion-based object detectors in the nuScenes dataset and achieves more robust results in adverse weather conditions and nighttime scenarios.

\end{abstract}

\section{Introduction}
\input{chapter/intro.tex}
\label{sec:intro}

\section{Related Work}
\input{chapter/related_work.tex}
\label{sec:related_work}

\section{Approach}
\input{chapter/approach.tex}

\label{sec:approach}


\section{Experiments}

\input{chapter/experiments.tex}
\label{sec:experiments}


\section{Conclusion}
\input{chapter/conclusion.tex}

\label{sec:conclusion}

\section{Acknowledgement}
Research leading to these results has received funding from the EU ECSEL Joint Undertaking under grant agreement n° 101007326 (project AI4CSM) and from the partners national funding authorities German Research Foundation (DFG, Deutsche Forschungsgemeinschaft) on behalf of German Ministry of Education and Research (BMBF).

\addtolength{\textheight}{-12cm}   







\bibliographystyle{IEEEtran}

\end{document}

%% file: chapter/intro.tex
Autonomous driving is a rapidly growing field that has the potential to revolutionize transportation. One of the key components of autonomous driving is object detection, which involves identifying and localizing objects in a scene \cite{survey2}. While vision sensors provide rich texture information about the environment, they are vulnerable to adverse weather conditions, affecting their image quality. On the other hand, millimeter Wave (mmWave) radar sensors are robust under all weather conditions and excel in estimating the distance and velocity of objects but provide less detailed information about the targets compared to cameras. This leads to a growing field in the research field of radar-camera fusion \cite{survey}.

The primary drawback of mmWave radar sensors utilized in public datasets, such as nuScenes \cite{nuscenes}, is the low elevation resolution, leading to the absence of height information in the radar points \cite{crf}. Furthermore, compared to lidar point clouds, radar point clouds are significantly sparser \cite{nuscenes}. However, they carry several properties of the detected object, such as distance information, relative velocity, and \ac{rcs}. Before fusing the radar data with visual images, preprocessing is required. The most common method for point cloud preprocessing involves projecting the points onto the perpendicular image plane \cite{rvnet, radar3d}, with various techniques employed to deal with sparsity. For instance, \cite{distant_vehicle, saf_fcos} generate small circles around each radar detection after the projection, while \cite{crf} utilizes height extension in a fixed value to address the lack of height information. The authors in \cite{rei, fpp} further spread the created vertical line in the azimuth direction according to the angle accuracy of the radar sensor. 

The performance of the network is directly impacted by the stage where fusion occurs. Integrating radar and image information at the input level requires high-quality data sources \cite{survey}. Therefore, many algorithms opt to fuse information at the feature level instead \cite{distant_vehicle, saf_fcos}. Some approaches attempt to fuse feature maps at various stages to improve performance \cite{rei, crf}. Furthermore, while multi-task learning has been extensively employed in lidar-camera fusion algorithms, such as in \cite{mono3d, multi}, it has yet to be thoroughly explored in the context of merging data from camera and radar. 

The preprocessing techniques mentioned above aim to increase the density of the radar point cloud, such as fixed-value height extension and circular mapping on the image plane. Nonetheless, these methods may also introduce erroneous information. Usually, the fusion of radar and image features using simple element-wise operations like addition, multiplication, or concatenation may not be optimal due to the inherent heterogeneity between the two modalities. In this study, we propose an \acrfull{ah} extension method that utilizes distance and \ac{rcs} information of radar points to improve input radar data quality. We also perform azimuth pixel extension resulting in an \acrfull{aue}. This helps to generate more reasonable radar inputs and gain a $2\%$ higher \ac{map} compared with the preprocessing in \cite{crf}. We also introduce two novel fusion blocks: the \acrfull{swfb} and the \acrfull{safb}, which allow for reweighting features from the camera and radar branches before fusion. 
We compare the effectiveness of our proposed fusion methods against commonly utilized fusion techniques such as concatenation, element-wise addition, and multiplication. Moreover, we also compare our approaches with recently popular techniques such as cross-attention \cite{attention} and \ac{cbam} \cite{cbam}. Our fusion blocks exhibit better performance, achieving approximately $1\%$ higher mAP. Furthermore, we put forth a multi-sensor, multi-task cross-modality fusion strategy designed to carry out both 2D object detection and free space segmentation tasks. This multi-task learning framework primarily aids in augmenting the radar latent space with additional information derived from the segmentation mask. The employment of multi-task learning results in a $1.5\%$ enhancement in mAP compared to conducting the detection task in isolation. We evaluate our approaches on the nuScenes dataset. The results show that our approach outperforms state-of-the-art radar-camera fusion algorithms with approximately $3\%$ higher mAP.

%% file: chapter/related_work.tex
This section provides an overview of the research on object detection for autonomous vehicles, categorized into single-modality and radar-camera fusion-based techniques.
\subsection{Single Modality Object Detection}
Early object detection solutions only focus on
camera images and are typically categorized into two
classes: one-stage and two-stage detectors.  The two-stage detectors, such as Faster-RCNN \cite{faster_rcnn} and Mask-RCNN \cite{mask-rcnn}, operate in two steps. In the first stage, they generate \ac{roi}s using a \ac{rpn} which pinpoints the most crucial parts of the image. The \ac{roi}s are then processed by a \ac{r-cnn} to refine the bounding boxes and perform object classification. Although two-stage detectors usually deliver better results, they often suffer from slow convergence and high computational demands. In contrast, one-stage detectors such as YOLO \cite{yolo}, SSD \cite{ssd} and RetinaNet \cite{retinanet} are designed to be faster since they replace the \ac{rpn} with pre-defined anchor boxes. This makes them a better choice for real-time object detection tasks. In this paper, we propose an architecture loosely inspired by the RetinaNet.


\subsection{Radar-Camera Fusion Object Detection}
 Most of the radar-camera fusion methods first preprocess the radar data to align with the camera image. Then the suitable fusion methods and stages are chosen properly to merge feature maps from different sensors and complete the object detection task.
\paragraph{Radar Data Preprocessing}
\label{subsec:related_preprocessing}
Point cloud-based radar data is usually projected onto the 2-D image plane by the camera intrinsic and both radar and camera extrinsic. Radar information of each point, such as velocity and \acs{rcs} value, are stored in different input channels. John et al. \cite{rvnet} use the projected pixel position for each radar detection directly, resulting in a highly sparse radar channel. To solve this problem, the authors in \cite{distant_vehicle, saf_fcos} project each radar detection as a solid circle and feed the information carried by the radar detection into extra radar channels. Differently, Nobis et al. \cite{crf} argue that projecting them as vertical lines is a better way to deal with the lack of height information. A fixed extension height value is used for all the radar detections. 
\cite{rei, fpp} follow the \acrfull{fh} extension strategy and further extend the vertical line in the horizontal direction. Given the measured azimuth angle with its accuracy, each vertical line can be then extended horizontally during the projection. This results in a variable-width extension, where the width depends on the angle accuracy and the distance. Meanwhile, the probability density curve is assumed to follow a Gaussian distribution, which is utilized to refine the extended pixels.

While the fixed height extension approach effectively helps to relocate the radar points, it may also mislead and bias the projection algorithm due to the neglect of other information. 
Specifically, as shown in Fig. \ref{fig:old_preprocessing}, most of the extended vertical lines are misaligned with the height of the corresponding objects. Additionally, this variable-width azimuth extension helps to solve the sparsity problem but leads to a higher computational requirement and dramatically increases the training time. Furthermore, azimuth extension only on \acs{rcs} channel results in the lack of other features in the extended pixels, which further confuses the algorithm. 
\paragraph{Fusion Methods and Stages}
\label{subsec:related_fusion}
The prevailing methods for input and feature-level fusion in existing literature mostly rely on concatenating \cite{rei,rvnet,crf} the two feature maps or using element-wise addition \cite{distant_vehicle} or multiplication \cite{saf_fcos} to merge them. However, these approaches overlook the significance of each pixel and the interdependence between the feature maps of different sensors. Chang et al.\cite{saf_fcos} introduce a spatial attention fusion block to solve this drawback. The radar feature map is only used to generate a weighting matrix utilizing convolutional layers with different kernel sizes. After applying this weighting matrix to the image feature map,  no more radar information is involved in further steps. Cross-attention \cite{attention} is another way to exchange and fuse information between two features, however, it dramatically increases the parameter size of the model.

It is well known that different fusion stages heavily influence the network performance. Most of the works \cite{distant_vehicle, saf_fcos, rvnet} fuse the features only once, so choosing the right fusion level is problematic. The work in \cite{crf} aims to solve this problem by concatenating feature maps at different levels. This enables the network to adjust its weight to the fusion points during training and makes the algorithm more robust.



%% file: chapter/approach.tex
In this section, we highlight the innovations of the proposed work. We first introduce our adaptive radar data preprocessing method. Then, we describe the architecture of our multi-task cross-modality fusion model. At last, we illustrate the two new fusion blocks in more detail.

\begin{figure}[htbp]
    \centering
    \begin{subfigure}[b]{.9\columnwidth}
    \centering
        \includegraphics[width=\textwidth]{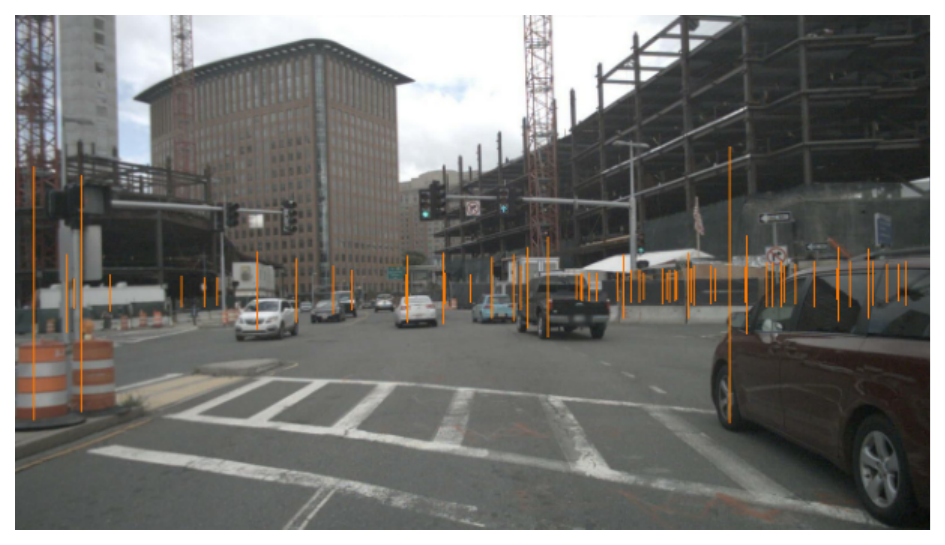}
        \caption{Preprocessing of radar points with fixed height value \cite{crf}.}
        \label{fig:old_preprocessing}
    \end{subfigure}
    \begin{subfigure}[b]{.9\columnwidth}
    \centering
        \includegraphics[width=\textwidth]{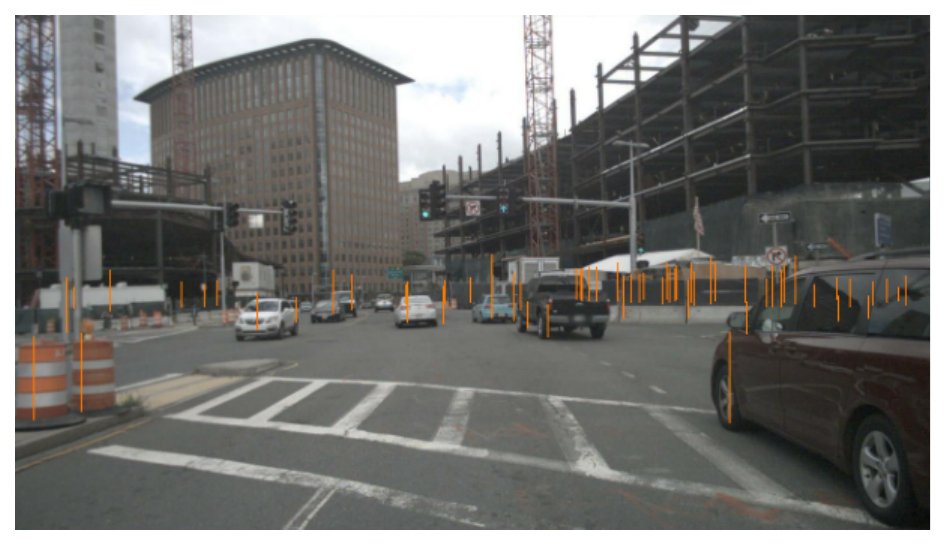}
        \caption{Distance- and \acs{rcs}-based height extension.}
        \label{fig:new_height}
    \end{subfigure}
\caption{Visualization of Height Extension Comparison.}
\vspace{-0.2cm}
\label{fig:height}
\end{figure}

\subsection{Radar Data Preprocessing}
\label{sec:approach_preprocess}
This section introduces the radar data preprocessing approaches that help to align radar points with image pixels.  
On the one hand, radars have the ability to accurately measure the distance and velocity of moving targets, even under adverse weather conditions. On the other hand, they struggle with poor elevation resolution, resulting in radar point clouds only containing 2D positional information. Therefore, in this work, we propose two novel approaches to preprocess the radar data and generate denser radar channels.
\paragraph{Height Extension}

As introduced in Sec. \ref{subsec:related_preprocessing}, the \acs{fh} extension has certain drawbacks, while the height should be estimated by other related factors. Generally, the farther the object is located, the less accurate the detection is since there are more obstacles and a decrease in signal power. Additionally, \acs{rcs} of the object has positive correlation with its size. 
Therefore, this study proposes a novel \acs{ah} approach to extend the height based on the distance and the \acs{rcs} value of the radar detection. For the $n^{th}$ radar point $p_{n}\in P$ in a dataset, its position is represented as $(x, y)$ with \acs{rcs} value $r_{n}$ where $r_{n} \in R$. The Euclidean distance, $d_{n}(x_{n}, y_{n}) \in D$, is calculated as $\sqrt{x_{n}^2 + y_{n}^2}$. 

Under the aforementioned conditions, we propose a new height estimation approach defined as follows:
\begin{equation}
\label{eq:height_est}
H_{n} = \max(H_{min}, \mathop{\min}_{d_{n} \in D, r_{n} \in R}(\alpha - \frac{d_{n}}{\mu_{d}}, \beta + \frac{r_{n}}{\mu_{r}}))
\end{equation}

\noindent where $\mu_{d}$ and $\mu_{r}$ are the mean distance and \acs{rcs} values over all radar points, respectively. 
In addition, parameters $\alpha$ and $\beta$ define two initial height estimations, which should be reduced or increased along with larger distance and \acs{rcs}, respectively. Then we select the larger one and clamp it with $H_{min}$ for the final estimation.
$H_{min}$, $\alpha$, $\beta$ need to be determined according to the height range of the objects in the dataset. 
 Further details are explained in Sec. \ref{subsec:implement}.

A visualization of the comparison between the FH and the AH extension is shown in Fig. \ref{fig:height}. As illustrated in Fig. \ref{fig:new_height}, the radar projections are better aligned with the objects after our \acs{ah} extension.

\paragraph{Azimuth Uncertainty}
In this work, we extend the projection to a fixed number $N$ of pixels on all radar channels, including distance, \acs{rcs} and velocities, namely \acrfull{ae}.
This method can significantly reduce computational complexity than variable-width extension without losing the key radar features in each pixel. For simplicity, we assume that the angle measurement follows a Gaussian distribution with the measured angle $\theta_{azi}$ of each radar point as mean and the azimuth angle accuracy $\theta_{acc}$ as standard deviation. It is then mapped onto the 2-D image plane and utilized to redistribute the \acs{rcs} value in extended pixels. This method is represented as \acrfull{aue} that adapts the pixels to the angle estimation uncertainty.

\begin{figure}[htbp]
    \centering
    \begin{subfigure}[b]{.9\columnwidth}
    \centering
        \includegraphics[width=\textwidth]{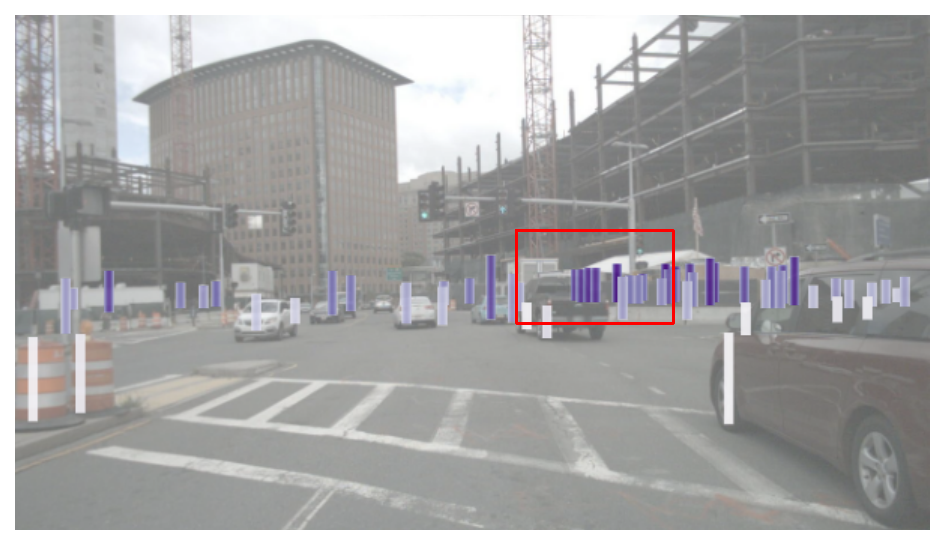}
        \caption{Camera image with the projected radar detections.
        }
        \label{fig:preprocessing_full}
    \end{subfigure}
    \begin{subfigure}[b]{.9\columnwidth}
    \centering
        \includegraphics[width=\textwidth]{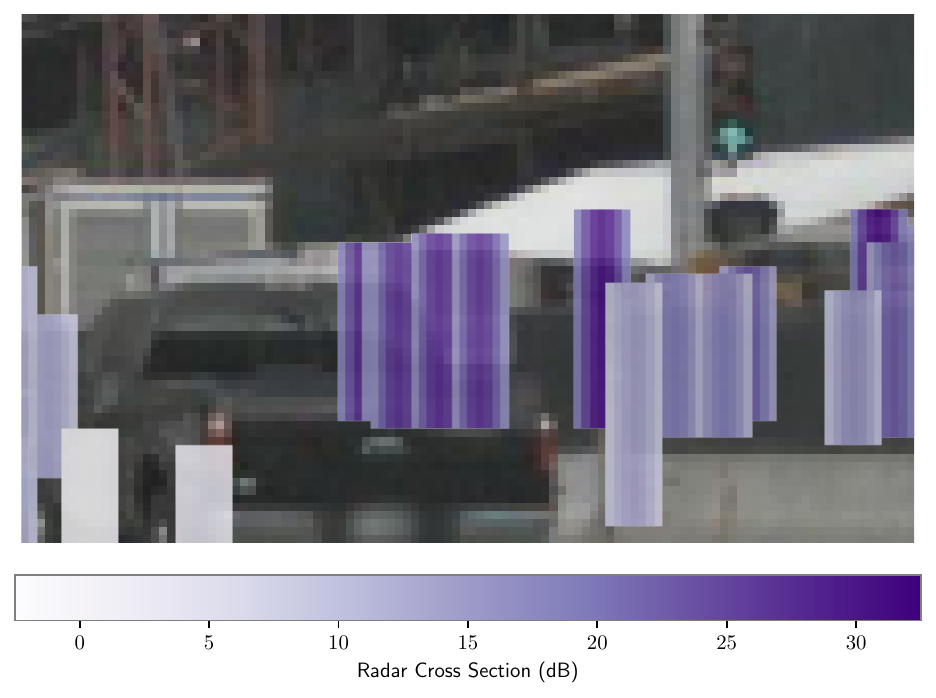}
        \caption{The closer look at the region highlighted in (a)}
        \label{fig:preprocessing_zoomin}
    \end{subfigure}
\caption{The visualization shows a combination of a camera image and the projection of the \acs{rcs} value of the radar data. (b) is a closer look at the region highlighted by the red bounding box in (a). The projected radar data is depicted in the form of rectangles. The \acs{rcs} values inside a rectangle follow a Gaussian distribution.}
\vspace{-0.4cm}  
\label{fig:preprocessing}
\end{figure}

A visualization of the proposed preprocessing method is shown in Fig. \ref{fig:preprocessing_full}. It displays the visual image combined with the projected radar data, specifically the \acs{rcs} channel. Fig. \ref{fig:preprocessing_zoomin} is a zoomed-in version of the red rectangle, demonstrating that the \acs{rcs} undergoes a change when spreading a single pixel width to multiple pixels.

The effectiveness of the proposed preprocessing methods is further analyzed in Sec. \ref{subsec:radar_analysis}.

\subsection{Model Architecture}
\label{sec:model}
Our MCAF-Net accomplishes two tasks, namely object detection and free space segmentation. 

For object detection, we utilize Retinanet, an one-stage detector, along with a pretrained \acf{cnn} backbone. In contrast, the radar feature extraction branch employs the \acs{cnn} backbone without pretrained weights. We acquire image feature maps ($C_{1}$-$C_{5}$) and radar feature maps ($R_{1}$-$R_{5}$) by gradually increasing the number of maps. Additionally, we apply an extra MaxPooling with strides of 2 to $R_{5}$ and $R_{6}$ to adjust the shape of the radar feature maps, resulting in $R_{6}$ and $R_{7}$, respectively. The \acs{swfb} processes the feature maps from the same level, such as $C_{3}$ and $R_{3}$, before they are fed into the \acrfull{fpn}. Finally, the classification and box regression subnets utilize the fused feature maps to generate a set of bounding boxes along with their corresponding classification results and 2D box coordinates.

The segmentation branch aims to enhance the quality of image and radar features, which guides the feature to focus more on the space with objects. To this end, we create a two-channel segmentation mask for each input image based on the bounding box annotations. The first channel has a value of 1 for free space and 0 for occupied space, while the free spaces in the second channel have a value of 0. This mask is created to locate object positions. The \acs{safb} is applied to $R_{5}$ and $C_{5}$, and the resulting features are passed through a decoder network consisting of blocks that employ a $3\times3$ transpose convolution and a $3\times3$ convolutional layer. The output has two channels and the same shape as the input image. 

\begin{figure*}
\begin{center}
\includegraphics[width=0.98\textwidth]{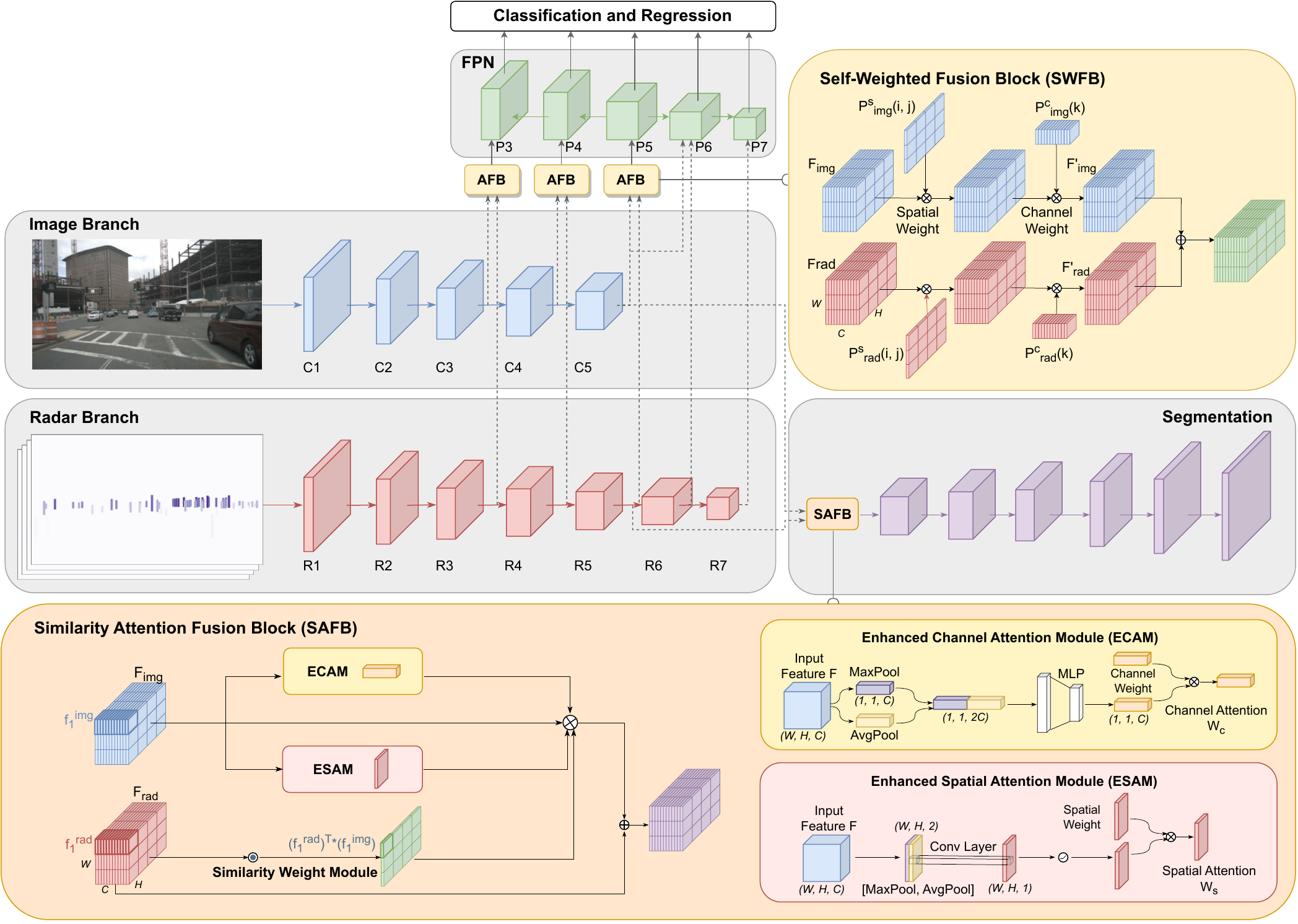}
\end{center}
   \caption{Model Architecture.}
\label{fig:model}
\vspace{-4mm}
\end{figure*}

Fig. \ref{fig:model} illustrates the network architecture, where the features are merged at various levels and before multiple branches. In the object detection task, the image and radar feature maps are fused at different stages of the \acs{fpn}. For the segmentation task, two features are merged using the \acs{safb}, and the result is fed as input to the segmentation subnet. This structure results in a multi-task multi-level cross-modality fusion.

The network is trained end-to-end with three losses: the classification loss $L_{cls}$, the regression loss $L_{reg}$, and the segmentation loss $L_{seg}$. We compute the $L_{cls}$ with the focal loss \cite{retinanet} to address the imbalance between foreground and background classes. $L_{reg}$ is calculated by a smooth $l1$ loss \cite{fast} on each dimension of the 2D bounding boxes $(x1, y1, x2, y2)$ and summed over positive samples for the regression head. For the segmentation branch, we use binary cross-entropy to compute $L_{seg}$ between the predicted masks and the ground truth masks.

\subsection{Fusion Block}
As mentioned in Sec. \ref{sec:model}, we introduce two new fusion blocks: the \acs{swfb} and \acs{safb}, which help to address the issue of immature fusion operations discussed in Sec. \ref{subsec:related_fusion}. These blocks allow for better correlation between the features of different sensors. Fig. \ref{fig:model} provides a visual representation of the blocks, and we will discuss them in more detail.

\paragraph{Self-Weighted Fusion Block}
Due to the intrinsic property of the radar sensor, there are no radar detections in most areas. Additionally, the camera image also contains redundant information. Thus, it is beneficial to redistribute the importance of each pixel before fusion. Let $F_{img}$ and $F_{rad}$ be two feature maps with the same shape $W \times H \times C$.  We introduce two trainable weighting maps, namely spatial weighting map $P^{s}$ and channel weighting map $P^{c}$, for each feature map. These weighting maps assist in redistributing the importance of each pixel and channel, respectively. The result coming from this proposed fusion block can be expressed as:
\begin{equation}
\begin{split}
    F_{fuse}(i,j,k) = F_{img}(i, j, k)\times P_{img}^{s}(i,j) \times P_{img}^{c}(k) \\
    + F_{rad}(i, j, k)\times P_{rad}^{s}(i,j) \times P_{rad}^{c}(k)
\end{split}
\end{equation}
\noindent where $i$, $j$, and $k$ represent the pixel location and channel index, respectively. 


\paragraph{\acl{safb}}
As illustrated in Fig. \ref{fig:preprocessing}, some of the radar detections are unrelated to any of the objects and may interfere with object detection. To address this issue, we introduce the segmentation branch that helps to filter the radar information and relocates the object position for features of both sensors. The radar feature map $F_{rad}$ is directly taken from the radar branch and serves as the input to the segmentation branch. Meanwhile, we propose three modules to refine the image feature map $F_{img}$, as shown in Fig. \ref{fig:model}. The Enhanced Channel and Spatial Attention Modules are inspired by \acs{cbam} \cite{cbam}, while we multiply the output of each module with an additional trainable weighting map, similar to \acs{swfb}. This results in a channel attention weight $W_{c}$ and a spatial attention weight $W_{s}$. These modules help to find the inter-relationship of the image feature map both channel and spatial-wise.
Furthermore, the Similarity Weight Module generates a weighting map $W_{sim}$ by calculating the similarity between $F_{rad}$ and $F_{img}$, which indicates the pixel relationships between the two feature maps. 
These three weights are multiplied with $F_{img}$ to scale the importance of each pixel, resulting in a modified image feature map $F'_{img}$. Then, the fused feature map $F_{fuse}$ is computed according to the Eq. \ref{eq:safb}:
\begin{equation}
\label{eq:safb}
    \begin{aligned}
       & \begin{split}
           F'_{img}(i, j, k) = & F_{img}(i,j,k)\times W_{sim}(i,j)\\
                        & \times W_{s}(i,j) \times W_{c}(k)
       \end{split}
        \\
       & F_{fuse}(i,j,k) = F_{rad}(i,j,k)+F'_{img}(i,j,k) 
    \end{aligned}
\end{equation}

By employing the \acs{safb}, we adjust the significance of individual pixels and channels in the image feature map based on its own characteristics and the interplay with the radar feature map. This yields a more comprehensive feature map, which aids in pinpointing the location of the objects and enhancing the radar features.

%% file: chapter/experiments.tex
In this section, we first explain the implementation details. Afterwards, we analyze the effectiveness of our radar preprocessing method. Then, we present the quantitative results with ablation studies of our proposed approaches. At last, we show the qualitative results compared with the CRF-Net.
\subsection{Dataset and Implementation Details}
\label{subsec:implement}
We use the nuScenes dataset, a state-of-the-art public dataset for autonomous driving, to train and evaluate our proposed network. It includes data from multiple sensors, including five radars, one lidar, and six cameras, and contains 3D bounding box annotations for 27 classes. As in \cite{crf}, we obtain 2D bounding boxes by projecting the 3D annotations onto the image plane and using the front camera and radar data. The dataset is split into 3:1:1 ratios for training, validation, and testing, including 20480, 6839, and 6830 frames, respectively. To ensure robustness to different weather conditions, we include a sufficient number of nightly and rainy scenes in the validation and test set. For most experiments, we group the 23 original object classes into seven high-level classes: car, truck, human, bicycle, motorcycle, bus, and trailer. We also conduct additional experiments using five classes, excluding bus and trailer, to compare with \cite{fpp}. In this study, we adopt the VGG16 \cite{vgg} architecture as the backbone in the experiments to align with the architectures utilized in previous studies \cite{crf,rei,fpp}. 

\begin{table*}[]
\caption{Quantitative Evaluation}
\label{table:experiment}
\resizebox{17.5cm}{!} 
{
\centering
\begin{tabular}{cl|ccccccc|ccc}
\multicolumn{1}{l}{}                                                                          &                       & Car & Truck & Human & Bicycle & Motorcycle & Bus & Trailer & mAP & Night mAP & Rain mAP \\ \hline
\multicolumn{1}{c|}{\multirow{4}{*}{\begin{tabular}[c]{@{}c@{}}Seven\\ Classes\end{tabular}}} & RetinaNet  \cite{retinanet}           & 53.33\%    &  25.00\%     & 40.25\%      &7.14\%         &  19.36\%          & 17.19\%    &  5.95\%       & 43.58\%     &  45.80\%         &  41.03\%        \\
\multicolumn{1}{c|}{}                                                                         & CRF-Net    \cite{crf}           & 53.81\%    & 19.09\%      & 41.28\%      &  14.28\%       & 25.16\%           & 12.28\%    &  14.80\%       & 43.83\%    &  46.12\%         & 41.04\%         \\
\multicolumn{1}{c|}{}                                                                         & REF-Net  \cite{rei}             &  --   &  --     & --      &  --       &    --        & --    &  --       & 44.76\%    &  47.64\%         &  41.23\%        \\ \cline{2-12} 
\multicolumn{1}{c|}{}                                                                         & MCAF-Net & 55.25\%    & 29.11\%      &  44.49\%     & 18.10\%        &  31.44\%          &42.94\%     &  24.74\%       & \textbf{47.70\%}    &  \textbf{49.77\%}         &  \textbf{44.91\%}        \\ \hline \hline
\multicolumn{1}{c|}{\multirow{4}{*}{\begin{tabular}[c]{@{}c@{}}Five\\ Classes\end{tabular}}}  & RetinaNet  \cite{retinanet}           & 53.64\%    & 24.28\%      & 41.19\%      &  9.85\%       &   9.07\%         & --    &  --       & 45.63\%    & 44.55\%          &   43.82\%       \\
\multicolumn{1}{c|}{}                                                                         & CRF-Net \cite{crf}              & 52.57\%    & 28.96\%      & 38.48\%      & 9.83\%        & 13.21\%           & --    &  --       &  44.75\%   & 42.20\%          &  42.92\%        \\
\multicolumn{1}{c|}{}                                                                         & UwRCS+FPP$^{\dag}$  \cite{fpp}           & 55.94\%    &  35.60\%     &  37.77\%     &   25.74\%      &  28.82\%         & --    &   --      & 46.73\%    &  50.53\%         &   --       \\ \cline{2-12} 
\multicolumn{1}{c|}{}                                                                         & MCAF-Net            &  54.73\%   &  31.99\%     &  45.54\%     & 18.91\%        &     33.00\%       &  --   & --        & \textbf{48.82\%}    &   \textbf{49.62\%}        &  \textbf{46.75\%}        \\ \hline
\end{tabular}
}
\begin{tablenotes}
      \small
      \item $^{\dag}$ These results come from \cite{fpp} based on a different data splitting (Mainly 6000 more training frames), which leads to the mAP metric being less comparable.
    \end{tablenotes}
\label{table:main_exp}
\end{table*}

We resize the images in the nuScenes dataset to $360 \times 640$ pixels to reduce the computational requirements. Afterwards, pixel values are further scaled to the interval $[-127.5, 127.5]$. For each image frame, a corresponding input of the radar branch is a four-channel image $(d, r, v_x, v_y)$, where $d$ indicates the distance, $r$ represents the \acs{rcs} value, and $v_x$ and $v_y$ denote the velocities along different axes. Pixels without corresponding radar points are filled with zero values in all radar channels. In overlapping entries, we retain the values of the radar points closest to the radar sensor. In addition, the parameters mentioned in Sec. \ref{sec:approach_preprocess} are chosen according to the nuScenes dataset. The set of distance values $D$, and absolute \acs{rcs} values $R$, are defined as follows: $D=\{d_{i} \mid d_{i} \in \mathbb{R}, d_{i} \in \left(0, 260\right)\}$, $R=\{r_{i} \mid r_{i}\in \mathbb{R},  r_{i} \in \left(-5, 53\right)\}$, respectively. The height of the objects being detected is in the range of one to five meters. Under these constraints, we choose the parameters for \acs{ah} extension as: $H_{min}=1$, $\alpha=6$ and $\beta=0.5$.  
For the azimuth extension, the ARS 408-21 radar sensor used in the nuScenes dataset has an azimuth angle accuracy of $\theta_{a}=\pm 0.3^{\circ}$ which is used as the variance of the Gaussian distribution. In addition, we spread the vertical line to three pixels in both left and right directions. 


We implement the networks using TensorFlow and train them on Nvidia\textsuperscript{\textregistered} Tesla\textsuperscript{\textregistered} P40 GPUs. The performance of object detection is evaluated based on the mAP metric, and we incorporate weights into the \acs{map} calculation to address the class imbalance, using an \ac{iou} threshold of 0.5. 
We implement training using the Adam optimizer over 40 epochs with a batch size of 12 for networks based on VGG. 
The learning rate set initially to $2e^{-4}$, is reduced by a factor of 0.75 whenever the optimization process hits a plateau.



\subsection{Radar Preprocessing Analysis}
\label{subsec:radar_analysis}
To assess the effectiveness of our proposed radar preprocessing methods, we conduct two experiments to compare the proposed methods with \acs{fh}. 

\begin{figure}[htbp]
\centering
     \includegraphics[width = 0.7\linewidth]{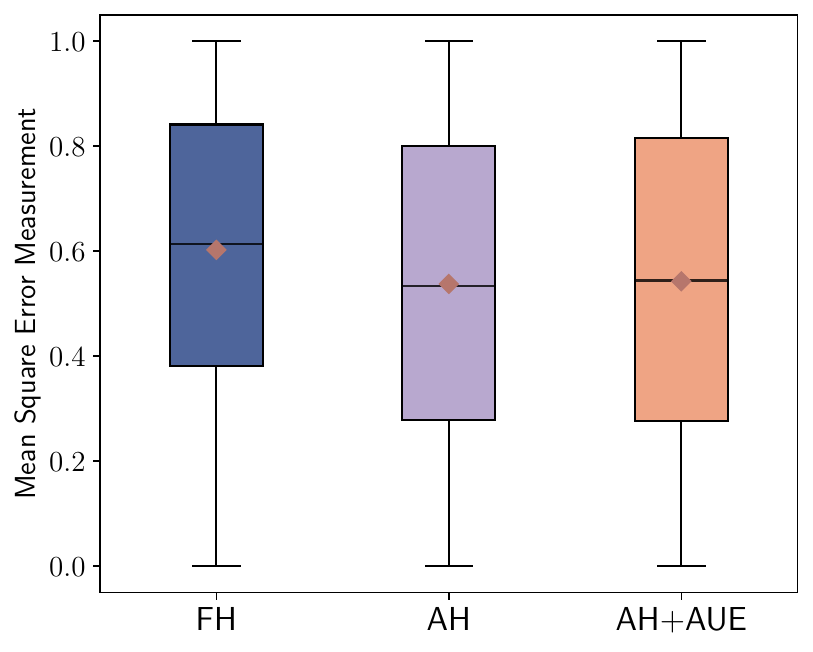}
     \caption{Projection Error Measurement on 2D plane.}
     \label{Fig: error}
     \vspace{-4mm}
 \end{figure}

Firstly, on the 2D image plane, we compute the ratio of projected radar points that fall outside the 2D ground truth bounding box. This ratio is quantified as a mean squared error (MSE) using the following formula:
\begin{equation}
    MSE^{k} = \frac{(n_{t}^{k}-n_{in}^{k})^{2}}{(n_{t}^{k})^{2}}
\end{equation}
For each frame $k$, after the radar points are projected onto the image plane, $n_{t}^{k}$ pixels carry radar information. Among these, $n_{in}^{k}$ pixels are located within the 2D ground truth bounding boxes. In Fig. \ref{Fig: error}, we visualize the distribution of MSE across the complete dataset for different preprocessing methodologies. Compared with \acs{fh}, it's evident that the use of \acs{ah} brings about a reduction in the average MSE. Moreover, with the use of \acs{ah}, 25\% of the dataset achieves an MSE of less than 0.3, a noteworthy decrease from the lower quartile mark of 0.4 attained by the \acs{fh} extension. 
By using \acs{aue}, we are able to augment the density of the radar projection channels without leading to an escalation in the MSE in comparison to \acs{ah}.

Next, we assess the error related to height within the 3D domain, taking the height of the 3D bounding box as the ground truth.  
We compute the height error $\delta h_{i}^{k}$ of the $i^{th}$ radar point in the $k^{th}$ frame as follows:
\begin{equation}
    \delta h_{i}^{k} = \left\{ \begin{array}{rcl}
            |H_{m}^{k} - h_{i}^{k}| & \mbox{if} & i \in B_{m}^{k} \\
            h_{i}^{k} & \mbox{if} & i \notin B^{k}
        \end{array} \right.
\end{equation}
If the radar point is inside $m^{th}$ bounding box $B_{m}^{k}$, the height error is defined as the absolute difference between the height of the bounding box $H_{m}^{k}$ and the measured extended height $h_{i}^{k}$. Otherwise, we directly take the estimated height as the error since the radar point does not belong to any objects.

\begin{figure}[htbp]
\centering
     \includegraphics[width = 0.7\linewidth]{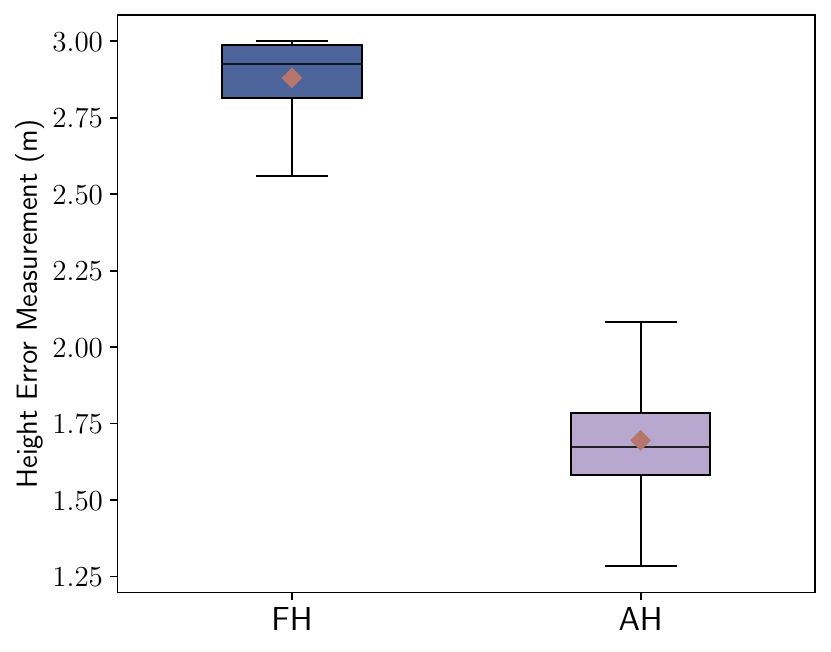}
     \caption{Height Error Measurement in 3D space.}
     \label{Fig: height_error}
     \vspace{-4mm}
 \end{figure}

The final height error $\Delta H^{k}$, is calculated by averaging the individual height errors $\delta h^{k}$ across all radar points. The distribution of the height error over the dataset is visualized in Fig. \ref{Fig: height_error}. By employing the \acs{ah} extension, we successfully reduce the average height error across the dataset from 2.9 meters to 1.7 meters. This remarkable improvement further underscores the effectiveness of the preprocessing method we propose.

\subsection{Quantitative Results}
We demonstrate the superior performance of our approach over state-of-the-art detectors in Table \ref{table:experiment}, achieving higher mAP across all classes and scenes. In our VGG-based MCAF-Net, the radar data is preprocessed utilizing AH and AUE. Additionally, SWFB and SAFB fusion approaches are applied to merge the radar and image feature maps. Our seven-class detector outperforms RetinaNet and CRF-Net by over 3\% mAP, with particularly significant gains of over 30\% for the Bus class. Notably, our network also indicates greater robustness in rainy-day scenarios, achieving a 3.68\% relative mAP increase compared to the REF-Net. In the case of the five-class detector, the CRF-Net does not manage to surpass its image-only baseline RetinaNet. However, our approach outperforms the RetinaNet by 3.19\% and achieves approximately 2\% higher mAP than the results reported in \cite{fpp} despite using less training data.



\begin{figure*}[t]
     \centering
     \begin{subfigure}[b]{0.31\textwidth}
         \centering
         \includegraphics[width=\textwidth]{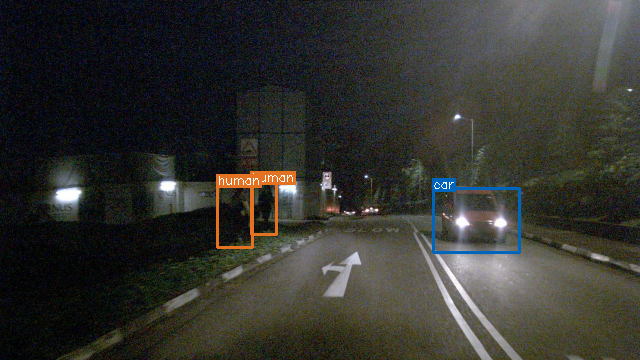}
         \label{fig:night-gt}
     \end{subfigure}
     \hfill
     \begin{subfigure}[b]{0.31\textwidth}
         \centering
         \includegraphics[width=\textwidth]{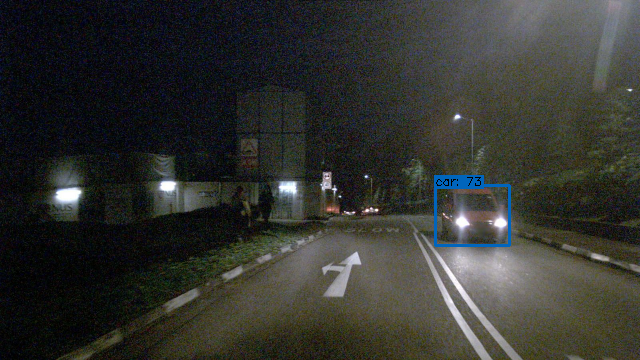}
         \label{fig:night-crf}
     \end{subfigure}
     \hfill
     \begin{subfigure}[b]{0.31\textwidth}
         \centering
         \includegraphics[width=\textwidth]{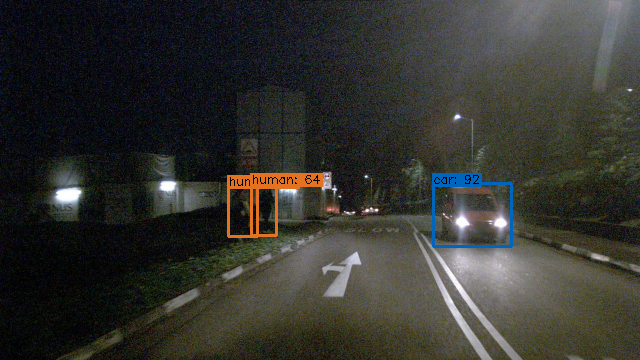}
         \label{fig:night-our}
     \end{subfigure}

     \begin{subfigure}[b]{0.31\textwidth}
         \centering
         \includegraphics[width=\textwidth]{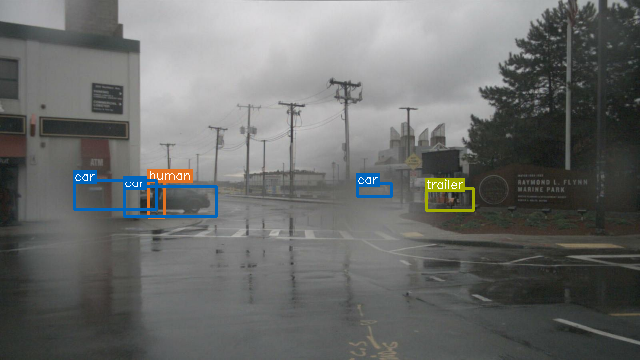}
         \caption{Ground Truth of 2D bounding boxes.}
         \label{fig:rain-gt}
     \end{subfigure}
     \hfill
     \begin{subfigure}[b]{0.31\textwidth}
         \centering
         \includegraphics[width=\textwidth]{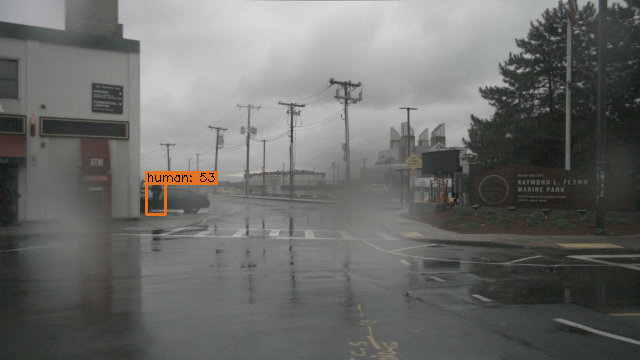}
         \caption{CRF-Net Detection.}
         \label{fig:rain-crf}
     \end{subfigure}
     \hfill
     \begin{subfigure}[b]{0.31\textwidth}
         \centering
         \includegraphics[width=\textwidth]{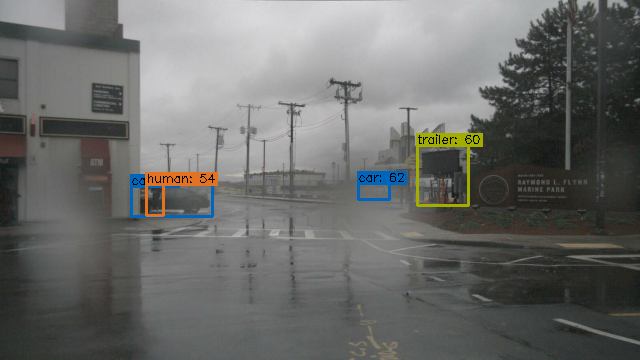}
         \caption{MCAF-Net Detection}
         \label{fig:rain-our}
     \end{subfigure}
        \caption{Qualitative comparison of detection results. Different classes are shown in different colors.}
        \label{fig:qualitative}
\vspace{-2mm}
\end{figure*}

\subsection{Ablation Study}
In this section, we conduct ablation studies on the nuScenes dataset to validate the effectiveness of our innovations. 
Firstly, we compare different preprocessing methods under the same network structure, as shown in Table \ref{table:preprocessing_comp}.

Next, we analyze the impact of multi-task learning. Previous works propose jointly training the network on object detection and distance estimation to improve proposal generation. However, our results in Table \ref{table:multi-task} show that incorporating distance estimation into our work results in declining network performance, despite adding over 2 million parameters. This is due to the increased difficulty for the network to prioritize object detection.

Finally, we compare the performance of our novel fusion blocks with other widely used fusion approaches, including concatenation, element-wise addition, multiplication, cross-attention, and \acs{cbam}-based fusion. In the \acs{cbam}-based fusion, the two feature maps undergo refinement by the CBAM and are subsequently combined using element-wise addition before being passed on to the downstream block. The results are presented in Table \ref{table:fusion_block}. The use of cross-attention or \acs{cbam} introduces over a million additional parameters, but surprisingly, this leads to a decrease in performance. This further reinforces that these two methods are not optimal for refining and fusing information for the camera and radar sensor. We ensure consistency in our approach by applying the same fusion strategy across all fusion stages.
It is important to note that the presented results are obtained from the model trained with seven high-level classes, as in \cite{crf}.

\begin{table}[]
\centering
\begin{tabular}{l||ccc}
\hline
Preprocessing & mAP & Night mAP & Rain mAP \\ \hline
FH  \cite{crf} & 45.79\%    &  47.83\%         & 42.83\%    \\
AH            & 46.77\%    &  48.53\%         &  43.53\%        \\
AH + AE       & 47.36\%    & 48.97\%          & 43.91\%         \\
AH + AUE      &  \textbf{47.70\%}    &  \textbf{49.77\%}         &  \textbf{44.91\%}          \\ \hline
\end{tabular}
\caption{Comparison of the benchmarked preprocessing methods. The performance of our proposed \acs{ah} extension method is compared with the \acs{fh} extension technique introduced in \cite{crf}. Moreover, we demonstrate that extending pixels in the azimuth direction can further enhance the network's performance.}
\label{table:preprocessing_comp}
\end{table}

\begin{table}[]
\centering
\begin{tabular}{l||ccc}
\hline
Model & mAP & Night mAP & Rain mAP \\ \hline
Baseline                                                                        &   46.28\%  &  47.12\%    &    42.07\%      \\
+ DS                                                           &    46.43\%   &   47.99\%      &  43.63\%        \\
+ Seg                                                                   & \textbf{47.70\%}    &  \textbf{49.77\%}         &  \textbf{44.91\%}          \\ 
+ DS + Seg &   47.28\%  &   49.20\%        &  44.32\%        \\ \hline
\end{tabular}
\caption{Ablation study of multi-task learning. ``DS'' and ``Seg'' denote distance estimation and segmentation tasks, respectively.}
\label{table:multi-task}
\end{table}

\begin{table}[]
\centering
\begin{tabular}{l||ccc}
\hline
Fusion & mAP & Night mAP & Rain mAP \\ \hline
Concatenation        &   46.74\%  &  47.82\%    &    43.61\%      \\
Addition            &    46.80\%   &   48.72\%      &  43.77\%        \\
Multiplication                       & 46.54\%    &  48.85\%         &   43.37\%       \\ 
Cross-Attention  & 45.61\%    &   42.72\%  &   43.20\% \\
CBAM-based fusion  & 44.95\%    &  46.06\%   & 40.95\%   \\
SWFB + SWFB &  47.23 \%  &   49.48  \%        & 44.53 \%        \\ 
SWFB + SAFB  & \textbf{47.70\%}    &  \textbf{49.77\%}         &  \textbf{44.91\%}        \\ \hline
\end{tabular}
\caption{Ablation study of fusion blocks. To demonstrate the effectiveness of our two novel fusion blocks, we conducted an experiment utilizing the \acs{swfb} at both fusion stages. Our results show that the \acs{swfb} combined with the \acs{safb} outperforms other fusion methods in all scenarios.}
\label{table:fusion_block}
\vspace{-0.4cm}
\end{table}

\subsection{Qualitative Results}
Two examples of the qualitative results are illustrated in Fig. \ref{fig:qualitative} by visualizing bounding boxes, classes and confidence probabilities. Compared with the CRF-Net, our proposed MCAF-Net conduces to a more robust detector, especially at night or under worse weather. The first row shows an example of the night scene. All objects are captured by our model, but the CRF-Net only detects the car and misses both two humans. This further indicates our algorithm takes better advantage of the radar information when the humans are almost invisible due to the poor lighting condition. Another example of the rainy scenario is represented in the second row, where the image is partially blurred caused by the raindrops. Meanwhile, the objects in this frame are small and some are even occluded, which increases the difficulty of detection. Thus, the fusion of radar data,  with an appropriate height extension, substantially improves the network performance. The effectiveness of our approach is proven as most of the objects are detected while the CRF-Net only has one human in its detection. 

%% file: chapter/conclusion.tex
This paper presents a multi-task detection model by fusing radar and camera data that jointly detects objects and segments free space. This multi-task learning helps the network learn better feature representations and concentrate on the position of objects. In addition, two new radar preprocessing techniques are proposed to handle the sparsity and uncertainty of the radar data, including an adaptive height extension and an azimuth uncertainty extension. These techniques help to generate denser radar input and better align the radar data with the objects on the image. 
Furthermore, the cross-modality fusion part is implemented through two new fusion blocks, the SWFB and SAFB. The SWFB is designed to redistribute the importance of pixels and channels for the radar and image feature maps before feeding them into the FPN. The SAFB aims more to reweight the image feature based on its relationship with the radar feature and its own inner-characteristics. Our approach outperforms RetinaNet by over 4\% mAP and achieves approximately 3\% mAP improvement compared to CRF-Net. In the future, we plan to extend our cross-modality fusion approach to jointly address the tasks of 3D object detection and depth completion.

%% file: main.bbl
\begin{thebibliography}{10}
\providecommand{\url}[1]{#1}
\csname url@rmstyle\endcsname
\providecommand{\newblock}{\relax}
\providecommand{\bibinfo}[2]{#2}
\providecommand\BIBentrySTDinterwordspacing{\spaceskip=0pt\relax}
\providecommand\BIBentryALTinterwordstretchfactor{4}
\providecommand\BIBentryALTinterwordspacing{\spaceskip=\fontdimen2\font plus
\BIBentryALTinterwordstretchfactor\fontdimen3\font minus
  \fontdimen4\font\relax}
\providecommand\BIBforeignlanguage[2]{{%
\expandafter\ifx\csname l@#1\endcsname\relax
\typeout{** WARNING: IEEEtran.bst: No hyphenation pattern has been}%
\typeout{** loaded for the language `#1'. Using the pattern for}%
\typeout{** the default language instead.}%
\else
\language=\csname l@#1\endcsname
\fi
#2}}

\bibitem{survey2}
J.~Fayyad, M.~A. Jaradat, D.~Gruyer, and H.~Najjaran, ``Deep learning sensor
  fusion for autonomous vehicle perception and localization: A review,''
  \emph{Sensors}, vol.~20, no.~15, p. 4220, 2020.

\bibitem{survey}
Y.~Zhou, L.~Liu, H.~Zhao, M.~L{\'o}pez-Ben{\'\i}tez, L.~Yu, and Y.~Yue,
  ``Towards deep radar perception for autonomous driving: Datasets, methods,
  and challenges,'' \emph{Sensors}, vol.~22, no.~11, p. 4208, 2022.

\bibitem{nuscenes}
H.~Caesar, V.~Bankiti, A.~H. Lang, S.~Vora, V.~E. Liong, Q.~Xu, A.~Krishnan,
  Y.~Pan, G.~Baldan, and O.~Beijbom, ``nuscenes: A multimodal dataset for
  autonomous driving,'' in \emph{Proceedings of the IEEE/CVF conference on
  computer vision and pattern recognition}, 2020, pp. 11\,621--11\,631.

\bibitem{crf}
F.~Nobis, M.~Geisslinger, M.~Weber, J.~Betz, and M.~Lienkamp, ``A deep
  learning-based radar and camera sensor fusion architecture for object
  detection,'' in \emph{2019 Sensor Data Fusion: Trends, Solutions,
  Applications (SDF)}.\hskip 1em plus 0.5em minus 0.4em\relax IEEE, 2019, pp.
  1--7.

\bibitem{rvnet}
V.~John and S.~Mita, ``Rvnet: Deep sensor fusion of monocular camera and radar
  for image-based obstacle detection in challenging environments,'' in
  \emph{Image and Video Technology: 9th Pacific-Rim Symposium, PSIVT 2019,
  Sydney, NSW, Australia, November 18--22, 2019, Proceedings 9}.\hskip 1em plus
  0.5em minus 0.4em\relax Springer, 2019, pp. 351--364.

\bibitem{radar3d}
M.~Meyer and G.~Kuschk, ``Deep learning based 3d object detection for
  automotive radar and camera,'' in \emph{2019 16th European Radar Conference
  (EuRAD)}.\hskip 1em plus 0.5em minus 0.4em\relax IEEE, 2019, pp. 133--136.

\bibitem{distant_vehicle}
S.~Chadwick, W.~Maddern, and P.~Newman, ``Distant vehicle detection using radar
  and vision,'' in \emph{2019 International Conference on Robotics and
  Automation (ICRA)}.\hskip 1em plus 0.5em minus 0.4em\relax IEEE, 2019, pp.
  8311--8317.

\bibitem{saf_fcos}
S.~Chang, Y.~Zhang, F.~Zhang, X.~Zhao, S.~Huang, Z.~Feng, and Z.~Wei, ``Spatial
  attention fusion for obstacle detection using mmwave radar and vision
  sensor,'' \emph{Sensors}, vol.~20, no.~4, p. 956, 2020.

\bibitem{rei}
Y.~Gu, S.~Meng, and K.~Shi, ``Radar-enhanced image fusion-based object
  detection for autonomous driving,'' in \emph{2022 IEEE International
  Conference on Signal Processing, Communications and Computing
  (ICSPCC)}.\hskip 1em plus 0.5em minus 0.4em\relax IEEE, 2022, pp. 1--6.

\bibitem{fpp}
L.~St{\"a}cker, P.~Heidenreich, J.~Rambach, and D.~Stricker, ``Fusion point
  pruning for optimized 2d object detection with radar-camera fusion,'' in
  \emph{Proceedings of the IEEE/CVF Winter Conference on Applications of
  Computer Vision}, 2022, pp. 3087--3094.

\bibitem{mono3d}
X.~Chen, K.~Kundu, Z.~Zhang, H.~Ma, S.~Fidler, and R.~Urtasun, ``Monocular 3d
  object detection for autonomous driving,'' in \emph{Proceedings of the IEEE
  conference on computer vision and pattern recognition}, 2016, pp. 2147--2156.

\bibitem{multi}
M.~Liang, B.~Yang, Y.~Chen, R.~Hu, and R.~Urtasun, ``Multi-task multi-sensor
  fusion for 3d object detection,'' in \emph{Proceedings of the IEEE/CVF
  Conference on Computer Vision and Pattern Recognition}, 2019, pp. 7345--7353.

\bibitem{attention}
A.~Vaswani, N.~Shazeer, N.~Parmar, J.~Uszkoreit, L.~Jones, A.~N. Gomez,
  {\L}.~Kaiser, and I.~Polosukhin, ``Attention is all you need,''
  \emph{Advances in neural information processing systems}, vol.~30, 2017.

\bibitem{cbam}
S.~Woo, J.~Park, J.-Y. Lee, and I.~S. Kweon, ``Cbam: Convolutional block
  attention module,'' in \emph{Proceedings of the European conference on
  computer vision (ECCV)}, 2018, pp. 3--19.

\bibitem{faster_rcnn}
S.~Ren, K.~He, R.~Girshick, and J.~Sun, ``Faster r-cnn: Towards real-time
  object detection with region proposal networks,'' \emph{Advances in neural
  information processing systems}, vol.~28, 2015.

\bibitem{mask-rcnn}
K.~He, G.~Gkioxari, P.~Doll{\'a}r, and R.~Girshick, ``Mask r-cnn,'' in
  \emph{Proceedings of the IEEE international conference on computer vision},
  2017, pp. 2961--2969.

\bibitem{yolo}
J.~Redmon and A.~Farhadi, ``Yolov3: An incremental improvement,'' \emph{arXiv
  preprint arXiv:1804.02767}, 2018.

\bibitem{ssd}
W.~Liu, D.~Anguelov, D.~Erhan, C.~Szegedy, S.~Reed, C.-Y. Fu, and A.~C. Berg,
  ``Ssd: Single shot multibox detector,'' in \emph{Computer Vision--ECCV 2016:
  14th European Conference, Amsterdam, The Netherlands, October 11--14, 2016,
  Proceedings, Part I 14}.\hskip 1em plus 0.5em minus 0.4em\relax Springer,
  2016, pp. 21--37.

\bibitem{retinanet}
T.-Y. Lin, P.~Goyal, R.~Girshick, K.~He, and P.~Doll{\'a}r, ``Focal loss for
  dense object detection,'' in \emph{Proceedings of the IEEE international
  conference on computer vision}, 2017, pp. 2980--2988.

\bibitem{fast}
R.~Girshick, ``Fast r-cnn,'' in \emph{Proceedings of the IEEE international
  conference on computer vision}, 2015, pp. 1440--1448.

\bibitem{vgg}
K.~Simonyan and A.~Zisserman, ``Very deep convolutional networks for
  large-scale image recognition,'' \emph{arXiv preprint arXiv:1409.1556}, 2014.

\end{thebibliography}
